\documentclass{article}

	\usepackage[preprint, nonatbib]{neurips_2020}

\usepackage{chngpage}
\usepackage[utf8]{inputenc} 
\usepackage[T1]{fontenc}    
\usepackage{hyperref}       
\usepackage{url}            
\usepackage{booktabs}       
\usepackage{amsfonts}       %
\usepackage{multirow}
\usepackage{amssymb}
\usepackage{amsmath}
\usepackage{nicefrac}       
\usepackage{microtype}      
\usepackage{neurips_2020}
\usepackage{upgreek}
\usepackage{graphicx}
\usepackage{subcaption}
\title{SEFR: A Fast Linear-Time Classifier for 
Ultra-Low Power Devices }

\author{%
Hamidreza Keshavarz\\
  Faculty of Electrical and Computer Engineering\\
  Tarbiat Modares University\\
  Tehran, Iran\\
  \texttt{keshavarz.h@modares.ac.ir} \\
  \And
  Mohammad Saniee Abadeh \\
  Faculty of Electrical and Computer Engineering\\
  Tarbiat Modares University\\
  Tehran, Iran\\
  \texttt{saniee@modares.ac.ir} \\
  \AND
  Reza Rawassizadeh\\
  Department of Computer Science, Metropolitan College \\
  Boston University \\
  \texttt{rezar@bu.edu} \\
}

\begin{document}

\maketitle

\begin{abstract}A fundamental challenge for running machine learning algorithms on battery-powered devices is the time and energy limitations, as these devices have constraints on resources. There are resource-efficient classifier algorithms that can run on these devices, but their accuracy is often sacrificed for resource efficiency. Here, we propose an ultra-low power classifier, SEFR, with linear time complexity, both in the training and the testing phases. SEFR is comparable to state-of-the-art classifiers in terms of classification accuracy, but it is \textbf{63 times faster} and \textbf{70 times more energy efficient} than the average of state-of-the-art and baseline classifiers on binary class datasets. The energy and memory consumption of SEFR is very insignificant, and it can even perform both train and test phases on microcontrollers. To our knowledge, this is the first multipurpose classification algorithm specifically designed to perform both training and testing on ultra-low power devices.
\end{abstract}
\section{Introduction}

There are many consumer electronic and medical devices that benefit from machine learning algorithms. From smartwatches to smart prosthetics, the devices are becoming more prevalent and smarter. However, the main part of learning, i.e. training, is usually done on a remote server and then, the model is deployed on the device. In other words, the data apparatus collects the data, sends them to a server via a network connection, receives the model from the server, and executes it. This process has two drawbacks. First, the connection may be lost or not available, which at some scenarios, such as medical devices, can become critical. Another disadvantage is that the privacy is not guaranteed, and many users may prefer not to send their personal data to a remote server or cloud. In addition, the server becomes the single point of failure, and if it crashes, the disconnected device with a trained model is prune to concept drift problem. In summary, reliability and privacy are two challenges of the existing edge, mobile and wearable technologies.

On-device machine learning can be a solution for these issues. However, mobile and wearable devices are usually constrained to low-power processors, and limited memory and battery capacity. Therefore, many machine learning algorithms cannot be run on these devices because of the time and memory constraints. Even algorithms that can be implemented on devices often suffer from low accuracies \cite{a}.

Resource limitations for training and testing, including runtime and energy utilization, is one of the fundamental challenges for supervised machine learning. A classifier typically gets training data, learns a model, and then classifies unlabeled data using the learned model. However, to mitigate the resource demand of classifiers, there are lazy classifiers in use, such as k-Nearest Neighbours (k-NN) \cite{knn}, in which no underlying model is trained. Although the k-NN classifier does not create a model, and thus, its training time complexity is $O(1)$, its testing phase can be computationally intensive. Therefore, methods such as KD-Tree \cite{n}, Voronoi Tessellation \cite{o}, and Locality Sensitive Hashing \cite{p} have been proposed to reduce the time complexity of its testing phase. 

Many accurate classifiers have been proposed in recent years, especially gradient boosting algorithms which show a significant superiority in their accuracy. Three prominent ones are CatBoost \cite{w}, LightGBM \cite{l}, and XGBoost \cite{k}. These methods have shown promising results, but they are not designed to be executed on low-power devices such as wearables. 

There are a few linear-time classification algorithms. The most notable one is Naïve Bayes \cite{g}, which has a reasonable runtime, but it works best when the features are independent. The accuracy of Naïve Bayes is a relatively poor compared to other algorithms on large datasets \cite{a}. Another linear time algorithm is Nearest Centroid Classification \cite{f}, in which samples that are too far from centroids may be classified incorrectly \cite{h}. Therefore, it is not widely in use, unlike previously listed classifiers.

While working with large amounts of data to create a model, many of the classification algorithms’ demand for runtime and energy grows exponentially. This becomes challenging when the number of features and/or training samples increases. To mitigate this challenge, distributed and parallel algorithms are used to train a model on large-scale and/or high-dimensional data, and architectures such as Spark \cite{spark} or hardware infrastructure such as Graphical Processing Unit (GPU) \cite{gpu} are in use. These approaches require powerful processors with multiple cores, large memory, and even then, they require significant time to train a model \cite{i}. Thus, it is common to perform their operations in batch mode. Furthermore, some algorithms, such as Support Vector Machines (SVM) \cite{svm}, are even slow on datasets which are not considered to be large-scale and/or high-dimensional \cite{j}. Therefore, due to the resource limitations, these algorithms are not used for battery powered devices, such as wearables, and it is impossible to train them on low power devices, such as microcontrollers. 

Several promising research works \cite{r, 55m, 105m} have used machine learning algorithms on edge devices. Besides, promising architectures such as Federated Learning \cite{federated} have been proposed to distribute the learning among low power clients and a powerful server. However, most of these efforts have been focused on testing phases, especially for ultra-low power devices such as microcontrollers \cite{q}, and still most of the training will be done on a remote machine. The computing power, the energy consumption, and the memory of these devices are very limited, and very few studies analyze both accuracy and complexity of machine learning algorithms for on-device learning \cite{m}. 

Existing efforts mainly focused on unsupervised learning which does not require model building \cite{xz}. Therefore, still an accurate and fast classification algorithm suitable for several settings that suffer from lack of computational capabilities, including ultra-low power devices, is missing. In this work, we aim to bridge this gap by proposing SEFR, a very fast classifier\footnote{SEFR is an acronym of Scalable, Efficient, and Fast classifieR.}.

We ran SEFR on running on five binary datasets, and in average, it consumes 2.14, 0.85, 0.47, 5.37, 6.82, 56.11, and 1.22 percent of energy required by LightGBM \cite{l}, XGBoost \cite{k}, CatBoost \cite{w}, Decision Tree, Random Forest, Naive Bayes, and SVM classifiers, respectively. Its accuracy and F1-score is comparable to computation-intensive classifiers, even ensemble learning classifiers. In average, its F1 score is 4.80, 4.75, 5.61, 5.67, and 4.49 percentage points lower than LightGBM, XGBoost, CatBoost, Random Forest, and SVM, and 2.22 and 5.51 percentage points higher than Decision Tree and Naive Bayes on binary class datasets, respectively. 

The SEFR classifier follows the explainable AI (xAI) paradigm and it gives weights to features. It starts by finding a hyperplane to separate two classes. It assigns a weight to each feature based on its correlation with each of two classes. This hyperplane partitions the space based on the two classes. Any test data point will be classified based on its location in the space. SEFR is designed for binary classification, but similar to other binary classifiers such as maximal margin classifier of SVM, it can handle multi-label classification by implementing one class versus other classes scheme. We ran SEFR on three multiclass datasets as well.

\section{Related Works}

Since we propose a classifier for on-device machine learning, this section consists of two subsections: on-device machine learning, and state-of-the-art classifiers.

\subsection{On-Device Machine Learning}
The biggest challenge for on-device learning is the resource constraints\cite{m}, and the machine learning algorithms should deal with this challenge. Considering resource constraints, on-device machine learning can be classified into two broad categories: resource-efficient inference, and resource-efficient training \cite{m}.

When applying resource-efficient inference, the data are usually gathered by the device and sent to a server, where a model is trained. Then, this model, either in a more compact form or in whole, is sent back to the device, and then the inference is accomplished on the device. One of the well-known paradigms is TinyML \cite{q}, in which a model in TensorFlow lite is transferred to an ultra low-power microcontroller. Capotondi et al. \cite{u} propose a convolutional neural network which is trained and transferred to a microcontroller. Another example is Bonsai \cite{r}, in which a tree-based algorithm is trained on a server, and the inference is done on a microcontroller. 

A common approach primarily used for mobile devices is Federated Learning \cite{v}, in which a server sends a global model across different devices,  and clients locally fine-tune their model based on their own training data, and send back their local models to the server. Afterward, the global model is updated based on the aggregations of local models. In the context of federated learning, training data are not sent to the server. Therefore, the accuracy of the learned model is lower in comparison to centralized models.

Another focus is on task specific algorithms that run on-device. When applying resource-efficient training, the focus is on creating new or optimizing existing algorithms such that they can run on resource-constrained devices \cite{m}. Some well-known machine learning algorithms that are optimized are Naive Bayes for parts-of-speech tagging \cite{s}, Gaussian Mixture Model \cite{t}, association rule mining \cite{xz} and a limited natural language processing \cite{yz}. Besides, several lightweight neural networks have also been proposed for training on mobiles \cite{55m,58m,119m}. However, despite promising results, most of these algorithms are application specific.

\subsection{State-of-the-art Classifiers}

Many promising classifiers have been proposed in the last decade, especially ensemble classifiers which show a significant impact. Three prominent ones are CatBoost \cite{w}, LightGBM \cite{l}, and XGBoost \cite{k}, which operate based on gradient boosting decision trees (GBDTs). LightGBM is a scalable algorithm that is efficient and accurate on large-scale and/or high-dimensional data. Its main advantage is that it selects a special set of features on the dataset. Another promising gradient boosting tree algorithm is XGBoost, which is sparsity-aware (can handle missing data), scalable, and accurate. XGBoost works by minimization of a regularized objective function, which consists of a convex loss function, and a penalty for the complexity of the model. CatBoost (Categorical Boosting)\cite{w} is a recent method that operates based on a novel solution based on gradient boosting, and avoids target leakage. Prokhorenkova et al. \cite{w} have shown that CatBoost outperforms previous GBDTs in accuracy. However, it is more energy-consuming and slower than XGBoost and LightGBM. The authors have proposed several ideas to improve the speed, such as using GPUs and tuning learning rate, but nevertheless, it generally consumes more energy and is slower than other state-of-the-art methods.

These methods have shown promising results, but they are not designed to be executed on low-power devices such as smartphones or wearables. Classic machine learning methods may need too much energy, or if not, their accuracy may suffer. The same issue exists for state-of-the-art ensemble classifiers, including XGBoost, CatBoost and LightGBM, as well. 

Therefore, we believe there is a need to devise new categories of multipurpose classification algorithms, designed for low-power devices, which cannot connect to a network and load the trained model, such as future medical devices that will be implanted inside the user's body.

\section{The SEFR Algorithm}

The SEFR algorithm is designed for binary classification. However, it could be extended to multi-class classications as well, the same way that other hyperplane based classifiers operate, such as the extension of SVM's maximal margin classifier to multi classes. Here, we explain the binary version of the SEFR algorithm, and then extend it to a multiclass version.

\subsection{Training and Inference}

SEFR labels one of the two classes as the negative class, and the other class as the positive class. This approach is inspired from work proposed by Keshavarz and Abadeh \cite{b}. The objective of SEFR algorithm is to find a hyperplane, by which the data are classified in space. A hyperplane can be shown as a matrix of weights and a bias, as it is shown in Equation \ref{eq1}.

\begin{equation}
w^Tx+b=0
\label{eq1}
\end{equation}

In \ref{eq1}, $w$ is the matrix of weights, $x$ is the input vector, and $b$ is the bias. If for the test input, $x_i$, $w^Tx_i+b \leqslant 0$, the record is classified as negative, and otherwise, it will be classified as positive.

In the context of the SEFR algorithm, we refer to the weights matrix as a feature dictionary, in which each of the features are assigned a score between $-1$ and $+1$. When a feature’s score is close to $-1$, it means that this feature is correlated with the negative class, and similarly, a score that is close to $+1$ means that the feature is correlated with the positive class. A feature score equal to 0 shows that the feature is irrelevant to the classification process.

The dataset that will be fed into the SEFR should be numerical, and the pre-processing phase should transform values in the range $[0, n]$ to avoid negative values. Here, we set $n$ to 1. Then, for each feature, its average values for positive and negative classes are calculated. To calculate them, we first define Equation \ref{eq2}, and then we will show the positive calculation in Equation \ref{eq3} and negative calculation in Equation \ref{eq4}.

\begin{equation}
p_i=\begin{cases}
0, & \text{if $L(x_i)$ = negative}.\\
1, & \text{if $L(x_i)$ = positive}.
\end{cases}
\label{eq2}
\end{equation}

\begin{equation}
\mu_\Delta(f_j)=\frac{1}{N_\Delta}\\\sum_{i=1}^{r}d_{i,j}p_i
\label{eq3}
\end{equation}

\begin{equation}
\mu_\nabla(f_j)=\frac{1}{N_\nabla}\\\sum_{i=1}^{r}d_{i,j}(1-p_i)
\label{eq4}
\end{equation}

$L(x_i)$ in Equation \ref{eq2} represents the label of the record $x_i$. $r$ in Equation \ref{eq3} and Equation \ref{eq4} presents the number of records, $i$ presents the record number, $j$ represents the feature number, $d_{i,j}$ is the value of feature $f_j$ in record $x_i$, and $N_\Delta$ and $N_\nabla$ represent the number of positive and negative instances in the training data, respectively.

After the average values for each feature has been calculated, then the weight of feature $j$, i.e. $w_j$,  is calculated as it has been shown in Equation \ref{eq5}.

\begin{equation}
w_j=\frac{\mu_\Delta(f_j)-\mu_\nabla(f_j)}{\mu_\Delta(f_j)+\mu_\nabla(f_j)+\epsilon}
\label{eq5}
\end{equation}

The feature score for feature $j$ shows its correlation with positive and negative classes. The feature score uses the normalization equation, and hence, its value is between $-1$ and $+1$. If it is close to $+1$, it shows that higher values in this feature have a positive impact on predicting the positive class. If it is close to $-1$, the higher values in this feature imply predicting the negative class. All scores are stored in a feature dictionary as weights. The $\epsilon$ is used to avoid division by zero. Next, for each record, a record score is calculated as it is shown in Equation \ref{eq6}.

\begin{equation}
e_i=w^Tx_i
\label{eq6}
\end{equation}

Now, each record $x_i$ is represented by a single numerical value, $e_i$. The higher $e$ show the tendency of their respective data points to the positive class, and the lower $e$ imply the tendency of the data points to be classified as the negative class. 
Afterward, we define a bias to distinguish positive and negative instances. This bias can be obtained using well-known classification algorithms; however, due to the premise of being linear-time, we propose to use the following approach to calculate this boundary.
First, the average scores for positive and negative classes are calculated as it is shown in Equation \ref{eq7} and Equation \ref{eq8}.

\begin{equation}
\uptau_\Delta=\frac{1}{N_\Delta}\\\sum_{i=1}^{r}e_{i}p_i
\label{eq7}
\end{equation}

\begin{equation}
\uptau_\nabla=\frac{1}{N_\nabla}\\\sum_{i=1}^{r}e_{i}(1-p_i)
\label{eq8}
\end{equation}

If the number of positive and negative instances are equal, the decision boundary will be the average of $\uptau_\nabla$ and $\uptau_\Delta$. However, unbalanced number of data points is a very common issue in real-world datasets, and thus, the weighted average is calculated for the decision boundary, as it is shown in Equation \ref{eq9}.

\begin{equation}
b=-\frac{\uptau_{\Delta}N_\nabla+\uptau_{\nabla}N_\Delta}{N_\nabla+N_\Delta}
\label{eq9}
\end{equation}

The resulted model can now be expressed in $w$ and $b$. For any data point $x_t$ in the test set, its score is calculated using $w^Tx_t+b$. If the resulting score is positive, the record will be classified as positive, and otherwise, it will be classified as negative.

\subsection{Time Complexity}
Assuming $n$ is the number of data points and $m$ is the number of features, the time complexity of SEFR classifier is $O(n.m)$ for both training and testing phases. Calculating $\mu_\Delta$ and $\mu_\nabla$ in Equation \ref{eq3} and Equation \ref{eq4} for all of the features is done in $O(n.m)$. Computing weights in Equation \ref{eq5} is done in $O(m)$. Then, the dataset should be traversed once more to calculate scores according to Equation \ref{eq6}, which is also in $O(n.m)$ time. Finding the bias consists of calculating two average values, $\uptau_\Delta$ and $\uptau_\nabla$ which are run in $O(n)$. Hence, the training time is $O(n.m)$. Testing of the model on one instance is done in $O(m)$, as it needs one dot product and one summation.

\subsection{Space Complexity}
The trained model which is used for testing consists of two components: (i) the weight matrix, which its size is equal to the number of features, and (ii) the bias, which is a single numerical value. Therefore, assuming $m$ is the number of features, the model just needs $m+1$ values. 

To calculate Equation \ref{eq3} and Equation \ref{eq4}, four variables are required (two for summations, one for the number of positive records, and one for the negative records). We keep the number of positive and negative records. For temporary storage of the values $\mu_\Delta(f_j)$ and $\mu_\nabla(f_j)$ in Equation \ref{eq3} and Equation \ref{eq4}, two variables are needed, which will be reused over features. After calculating them for each feature, the weight of that feature is calculated using Equation \ref{eq5}, and stored.

For temporary storage of values of Equation \ref{eq6}, one variable is needed that will be reused for each record in Equation \ref{eq7} and Equation \ref{eq8}. For Equation \ref{eq7} and Equation \ref{eq8}, two variables are needed to keep sums of positive and negative records, as we have numbers of positive and negative records before. These two variables can be reused from the previous steps of the algorithm, as they are no longer needed for those parts. After these steps, the bias is computed as in Equation \ref{eq9}.

The variables used in the \textit{For} loops can be reused. Hence, assuming we have $m$ features for training, the space complexity of training is $O(m)$. 
 
For testing phase on a test example, only one variable is needed to calculate the dot product of the weight matrix and the test record, and then adding the bias. Hence, the testing phase requires just one variable, and its space complexity is $O(1)$. 

\normalsize
\subsection{Multiclass Classification}
We extend SEFR to multiclass classification by using the one-against-all scheme. Assume $c$ is the number of classifiers that are trained, and it is equal to the number of classes. For classifier number $i$, the positive class is all classes minus $i$, and the negative class is the class $i$. Then, $w^Tx+b$ is computed for each classifier, and the assigned class will be selected according to the minimum value of $w^Tx+b$. As there are $c$ classifiers to be trained, the time complexity for training will be $O(m.n.c)$, and the space complexity will be $O(m.c)$.
\section{Experimental Evaluations}
In this section, the SEFR algorithm is compared to state-of-the-art and baseline classifications algorithms. The comparisons are based on accuracy, F-measure, execution time, and energy consumption. Our results show that the accuracy of the SEFR method is comparable to state-of-the-art classifiers, and its energy utilization is significantly lower than all state-of-the-art algorithms. Next, the scalability of SEFR and other methods will be analyzed. Since SEFR follows the explainable AI paradigm, in the following experiment, weights of the algorithm on image data are visualized to demonstrate its explainability as well. Finally, the execution time of SEFR on different partitions of a dataset on an Arduino Uno \footnote{https://store.arduino.cc/usa/arduino-uno-rev3} microcontroller will be presented.

Our experiments were conducted on the Ubuntu operating sytsem version 20.04, with an Intel Core i7-8750H CPU @ 2.20 GHz and 16 GBs of RAM. The programs on benchmark datasets were written in Python 3.8.2. Only one core of the CPU was used \footnote{The code is available at https://github.com/sefr-classifier/sefr}.

\subsection{Datasets}
 We evaluate the SEFR algorithm  on five benchmark, and high-dimensional datasets, namely GLI\_85\footnotemark, SMK\_CAN\_187\footnotemark[\value{footnote}], BASEHOCK\footnotemark[\value{footnote}], Gisette\footnotemark[\value{footnote}], 
 \footnotetext{https://jundongl.github.io/scikit-feature/datasets.html}
 and Sonar\footnote{http://networkrepository.com/sonar.php}. To perform  multiclass classification, we evaluate SEFR on three datasets, namely CNAE-9\footnote{https://archive.ics.uci.edu/ml/datasets/CNAE-9}, Waveform 5000\footnote{https://datahub.io/machine-learning/waveform-5000}, and Semeion\footnote{https://archive.ics.uci.edu/ml/datasets/semeion+handwritten+digit}. Table \ref{tab:table1} reports about quantitative information of these datasets. These datasets include medical, text, digit recognition, wave, and sonar signal information. In the pre-processing phase, they are transformed to range $[0, 1]$ using min-max normalization before running algorithms.
 \begin{table*}[htb]
  \caption{Quantitative descriptions of datasets used for experiments.}
  
  \centering
  
  \begin{tabular}{lccc}
    \toprule
 Dataset & Number of Records & Number of Features & Number of classes\\ [0.5ex] 
 
\hline
   \midrule
 GLI\_85 & 85 & 22283 & 2 \\
 SMK\_CAN\_187 & 187 & 19993 & 2 \\
 BASEHOCK & 1993 & 4862 & 2 \\
 Gisette & 7000 & 5000 & 2\\
  Sonar & 208 & 60 & 2\\
  CNAE-9 & 1080 & 856 & 9 \\
  Waveform 5000 & 5000 & 40 & 3 \\
  Semeion & 1593 & 256 & 10 \\
  
    \bottomrule
  \end{tabular}
  \label{tab:table1}
\end{table*}

\tabcolsep=0.11cm

 \begin{table*}[!h]

  \fontsize{8}{10}\selectfont
\hspace{-10mm}

     \caption{Accuracy and F-Measure comparison between different binary class datasets and algorithms.}
     \centering
\begin{tabular}{lcccccccccc}
              & \multicolumn{2}{c}{GLI\_85} & \multicolumn{2}{c}{SMK\_CAN\_187} & \multicolumn{2}{c}{BASEHOCK} & \multicolumn{2}{c}{Gisette} & \multicolumn{2}{c}{Sonar} \\ \hline
Method        & Accuracy       & F1         & Accuracy          & F1            & Accuracy       & F1          & Accuracy       & F1         & Accuracy      & F1        \\ \hline
LightGBM      & 79.58          & 69.71      & 67.40             & 65.15         & 96.94          & 96.93       & 97.93          & 97.93      & 88.38         & 87.62     \\
XGBoost       & 85.56          & 75.91      & 64.65             & 62.59         & 96.44          & 96.43       & 97.94          & 97.94      & 85.07         & 84.24     \\
CatBoost      & 87.92          & 78.96      & 68.30             & 65.80         & 95.53          & 95.52       & 97.06          & 97.05      & 84.64         & 84.07     \\
Decision Tree & 81.11          & 70.12      & 55.50             & 53.44         & 94.78          & 94.72       & 93.26          & 93.25      & 72.02         & 70.74     \\
Random Forest & 87.08          & 79.16      & 66.20             & 63.66         & 97.74          & 97.73       & 97.36          & 97.35      & 84.62         & 83.80     \\
Naive Bayes   & 80.00          & 72.29      & 58.47             & 56.21         & 95.79          & 95.78       & 73.50          & 71.82      & 70.74         & 69.72     \\
SVM           & 89.31          & 80.30      & 68.98             & 66.96         & 96.14          & 96.12       & 97.21          & 97.21      & 76.43         & 75.24     \\
SEFR          & 85.69          & 79.15      & 63.57             & 61.95         & 94.68          & 94.83       & 88.18          & 88.16      & 70.17         & 69.27    
\end{tabular}
\label{tab:table2}
\end{table*}
\tabcolsep=0.11cm
\tabcolsep=0.11cm

 \begin{table*}[!h]

  \fontsize{8}{10}\selectfont
\hspace{-10mm}

     \caption{Accuracy and F-Measure comparison between different multiclass  datasets and algorithms.}
     \centering
\begin{tabular}{lcccccccccc}
              & \multicolumn{2}{c}{CNAE-9} & \multicolumn{2}{c}{Waveform 5000} & \multicolumn{2}{c}{Semeion} \\ \hline
Method        & Accuracy       & F1         & Accuracy          & F1            & Accuracy       & F1                  \\ \hline
LightGBM      & 86.02          & 85.82      & 85.66             & 85.62         & 92.53          & 92.31            \\
XGBoost       & 91.39          & 91.09      & 85.02             & 84.97         & 92.28          & 92.07            \\
CatBoost      & 92.78          & 92.65      & 84.14             & 84.10         & 91.27          & 90.92     \\
Decision Tree & 87.22          & 86.99      & 73.70             & 73.67         & 75.70          & 74.94            \\
Random Forest & 92.41          & 92.27      & 85.32             & 85.25         & 93.97          & 93.69     \\
Naive Bayes   & 87.50          & 87.09      & 79.98             & 78.80         & 78.47          & 78.79       \\
SVM           & 95.65          & 95.51      & 86.86             & 86.83         & 88.14          & 87.84            \\
SEFR          & 90.83          & 90.85      & 84.08             & 83.73         & 83.49          & 83.54
\end{tabular}
\label{tab:table2m}
\end{table*}

\subsection{Accuracy}
We compare the results of our algorithm to three state-of-the-art classifiers, LightGBM, XGBoost, and CatBoost, and four benchmark classifiers, Decision Tree, Random Forest, Naïve Bayes, and SVM with the 10-fold cross-validation mechanism. The results, reported in Table \ref{tab:table2}, present accuracy and F-measure. The algorithms were used with default parameters; SVM was run with a linear kernel, and to have a fair comparison, CatBoost was run with 50 iterations. The default number of iterations in CatBoost is 1000, but it would take much more time and energy.

As the results show, SEFR is comparable to state-of-the-art and baseline methods. In terms of F-Measure, SEFR outperforms Naive Bayes on 3 out of 5 datasets. 

Table \ref{tab:table2m} shows the accuracy and F-measure on multiclass datasets.

\subsection{Runtime and Energy Consumption}
In this subsection, the runtime and energy consumption of each algorithm (in 10-fold cross-validation) are reported. The power consumption is calculated using the powerstat program \footnote{http://manpages.ubuntu.com/manpages/xenial/man8/powerstat.8.html}, which calculates the energy that the device utilizes in watts, while running programs. The idle power consumption is subtracted from the reported power consumption during running the algorithms. To calculate the energy (joules), we multiply the power consumption (in watts) to the runtime (in seconds) of the algorithm. We ran each algorithm ten times and report the average numbers.
The results show the average F-measure of SEFR is 5.51 and 2.22 percentage points better than Naive Bayes and Decision Tree, respectively.   

Table \ref{tab:table3} and Table \ref{tab:table3m} provide a comparison of the execution time (in seconds) and energy utilization (in joules) between each algorithm on binary and multiclass datasets, respectively. 

\begin{table*}[!h]
\begin{adjustwidth}{-.5in}{-.5in}  
\centering

 \fontsize{8}{10}\selectfont
 \hspace{-10mm}
\caption{Power and energy consumption comparison between different binary class datasets and algorithms.}

\begin{tabular}{lcccccccccc}
              & \multicolumn{2}{c}{GLI\_85} & \multicolumn{2}{c}{SMK\_CAN\_187} & \multicolumn{2}{c}{BASEHOCK} & \multicolumn{2}{c}{Gisette}                                      & \multicolumn{2}{c}{Sonar}                                        \\ \hline
Method        & Runtime (s)   & Energy (J)  & Runtime (s)      & Energy (J)     & Runtime (s)   & Energy (J)   & \multicolumn{1}{l}{Runtime (s)} & \multicolumn{1}{l}{Energy (J)} & \multicolumn{1}{l}{Runtime (s)} & \multicolumn{1}{l}{Energy (J)} \\ \hline
LightGBM      & 18.01         & 598.23      & 42.78            & 1538.76        & 6.33         & 205.69       & 155.75                          & 6061.63                        & 0.50                            & 15.60                         \\
XGBoost       & 10.13         & 418.35      & 25.92            & 1211.88        & 78.61         & 3753.48      & 317.46                          & 15739.57                       & 0.38                            & 11.82                          \\
CatBoost      & 324.30        & 12764.45    & 469.63           & 19846.56       & 18.22         & 498.76       & 125.99                          & 5406.32                        & 4.22                            & 66.04                          \\
Decision Tree & 3.59          & 118.48       & 10.12             & 325.29         & 7.11          & 219.29       & 85.55                           & 2698.94                        & 0.08                            & 2.54                           \\
Random Forest & 3.75          & 121.61       & 6.65             & 213.93         & 10.82         & 346.78       & 59.67                           & 1921.37                        & 1.40                            & 43.99                          \\
Naive Bayes   & 1.77          & 59.15       & 1.79             & 60.31          & 1.54          & 49.28        & 4.69                            & 151.78                         & 0.05                            & 1.29                           \\
SVM           & 3.09          & 93.94       & 8.76             & 241.65         & 83.00         & 2105.68      & 467.02                          & 12408.62                       & 0.06                            & 1.70                           \\
SEFR          & 0.36          & 12.98       & 0.56             & 19.88          & 1.35          & 46.58        & 3.05                           & 98.66                         & 0.07                            & 2.46                         
\end{tabular}
\label{tab:table3}
\end{adjustwidth}
\end{table*}

\tabcolsep=0.11cm
\begin{table*}[!h]
\caption{Power and energy consumption comparison between different multiclass datasets and algorithms.}
\centering
 \fontsize{8}{10}\selectfont
\hspace{-10mm}

\begin{tabular}{lcccccccccc}
              & \multicolumn{2}{c}{CNAE-9} & \multicolumn{2}{c}{Waveform 5000} & \multicolumn{2}{c}{Semeion}                                 \\ \hline
Method        & Runtime (s)   & Energy (J)  & Runtime (s)      & Energy (J)     & Runtime (s)   & Energy (J)\\   \hline
LightGBM      & 10.30         & 295.30      & 9.97            & 351.14        & 20.17         & 740.44          \\
XGBoost       & 51.77         & 2326.02      & 14.90            & 660.07        & 21.18         & 945.69                  \\
CatBoost      & 7.36        & 159.86    & 8.80           & 226.42       & 17.87         & 704.01           \\
Decision Tree & 0.41          & 12.23       & 2.18             & 67.69         & 1.05          & 31.52             \\
Random Forest & 1.87          & 58.05       & 16.25             & 524.71         & 4.76         & 155.89                     \\
Naive Bayes   & 0.30          & 10.46       & 0.07             & 2.22          & 0.16          & 4.84             \\
SVM           & 19.57          & 478.68       & 6.26             & 177.91         & 10.45         & 271.39             \\
SEFR          & 20.61          & 680.54       & 0.48             & 17.27          & 1.65          & 58.61                      
\end{tabular}
\label{tab:table3m}
\end{table*}

\subsection{Scalability}
\begin{figure*}[h!]
  \centering

    \includegraphics[width=\linewidth]
    {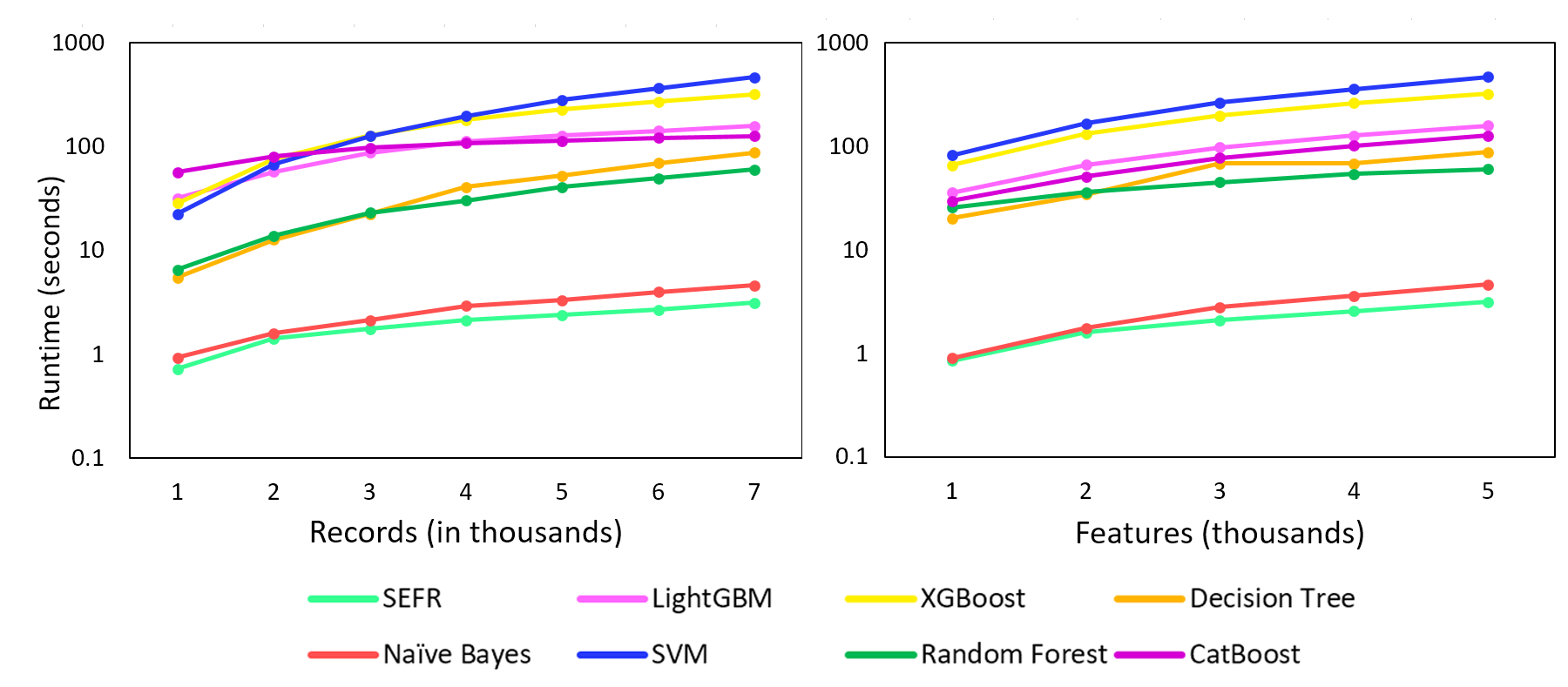}
    \caption{Runtime of the algorithms on the Gisette dataset with varying number of records and  features}

\label{fig1}
\end{figure*}

This section demonstrates the scalability of the SEFR algorithm in comparison to baseline and state-of-the-art algorithms. To this end, we ran the six algorithms 10 times on the Gisette dataset which has 7,000 records and 5,000 features. In six runs, we sampled the dataset such that 1,000, 2,000, …, 6,000 records are selected (with all of the features), and in four runs, we took 1,000, 2,000, 3,000, and 4,000 features, while maintaining the number of records. The runtime results are presented in Figure \ref{fig1}. As it can be seen in this figure, SEFR is the most scalable algorithm. 

\subsection{Explainability}
\begin{figure*}[h!]
  \centering

    \includegraphics[width=\linewidth]
    {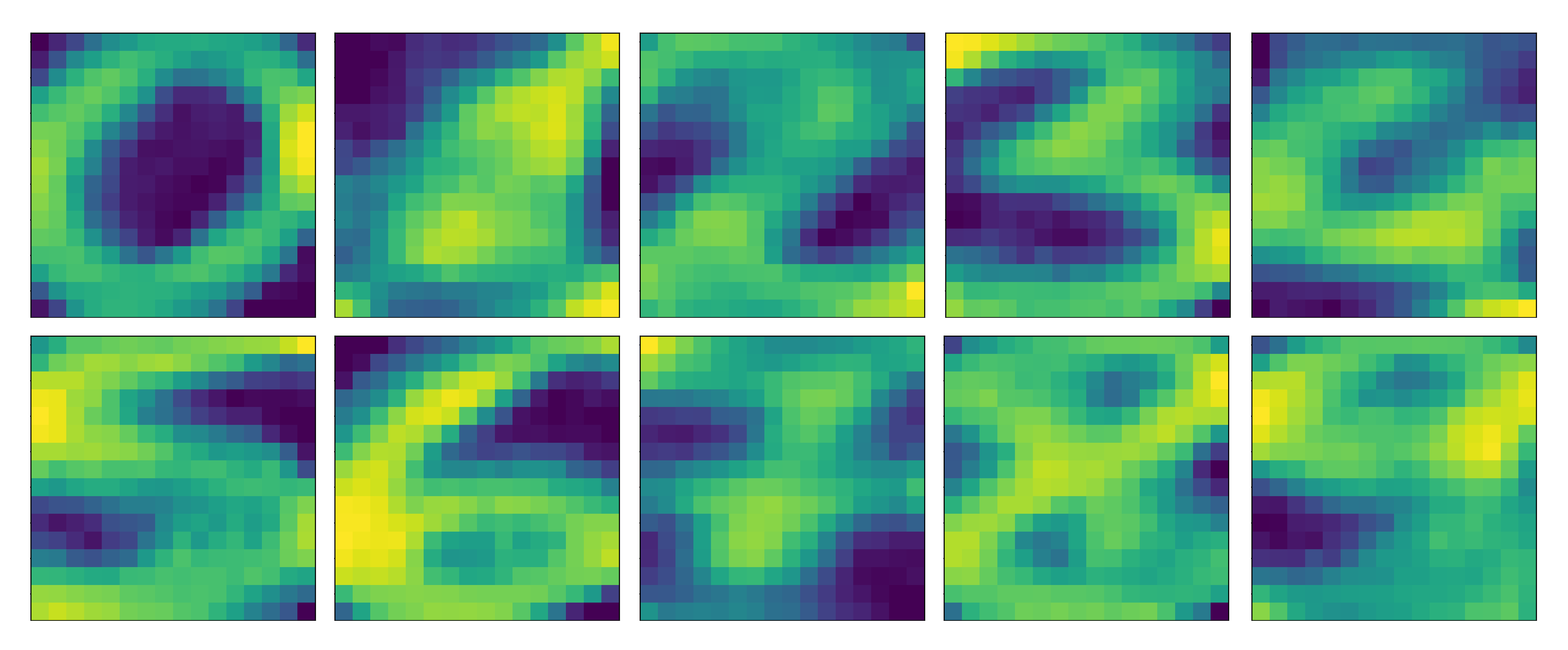}
    \caption{Visualization of SEFR weights on the Semeion digits dataset}

\label{fig2}
\end{figure*}

The SEFR method follows the xAI paradigm, and thus its weights and biases can be read and interpreted. Weight of a feature shows its correlation with positive and negative classes. As an example, we apply SEFR training on the Semeion dataset, which consists of images of handwritten digits, and has 256 features which represent the 16x16 pixels of each image. The weights for each class, i.e. digit, and each feature, i.e. pixel, are calculated, and are in the range $[-1,1]$. The number of weights is equal to the number of features, and when the weights are visualized as 16x16 images, they show the learned patterns for each digit in the one-against-all scheme. These patterns can be seen in Figure \ref{fig2}, and are readable by human. For each class (digit) a classifier is built, and the weights are shown. It can be seen that SEFR highlights the edges on each image with a spectrum of yellow to green color.

\subsection{Running of SEFR on Arduino}
To show the performance of SEFR, it has been implemented on an Arduino Uno Rev3 microcontroller. Its processor is ATMEGA328P, which runs in 16 MHz, its power consumption is about 0.30 watts, and has 2 KB of SRAM. The code was written in the C language to be implemented on the microcontroller, and it uses significantly less power than the Python implementation. Since the flash memory of the device has just 32 KB capacity, each dataset value is converted to a single unsigned byte, and then we transfer them into the device flash memory. 

To train and test on Arduino Uno, we partitioned the Gisette dataset, and created sub-datasets with lower number of records and features. For training, sub-datasets with 25, 50, 75, and 100 records, and 25, 50, 75, and 100 features were used, and the runtime of training on each sub-dataset is reported in Table \ref{tab:table4}. The trained model was applied on sub-datasets with 25, 50, 75, and 100 records, and 25, 50, 75, and 100 features. The runtime results are reported in Table \ref{tab:table5}. Since testing on each sub-dataset took less than 1 millisecond to complete, the model was applied 100 times on each of the sub-datasets, and the average runtime is reported.

\begin{table}[htb]

\caption{Training runtime of SEFR  with different number of records and features, in milliseconds.}
  \centering
\begin{tabular}{llllll}
                                              &                                   & \multicolumn{4}{c}{Features}                           \\ \cline{3-6} 
                                              &                                   & 25 & 50 & 75 & 100 \\ \cline{3-6} 
\multicolumn{1}{l|}{} & \multicolumn{1}{l|}{25}  & 13          & 27          & 41          & 55           \\
\multicolumn{1}{l|}{}                         & \multicolumn{1}{l|}{50}  & 26          & 51          & 76          & 102          \\
\multicolumn{1}{l|}{Records}                         & \multicolumn{1}{l|}{75}  & 37          & 74          & 112         & 150          \\
\multicolumn{1}{l|}{}                         & \multicolumn{1}{l|}{100} & 48          & 97          & 146         & 195
\label{tab:table4}
\end{tabular}
\end{table}


\begin{table}[!h]
  \centering
\caption{Testing runtime of SEFR with different number of records and features, in milliseconds}
\begin{tabular}{llllll}
                                              &                                   & \multicolumn{4}{c}{Features}                           \\ \cline{3-6} 
                                              &                                   & 25 & 50 & 75 & 100 \\ \cline{3-6} 
\multicolumn{1}{l|}{} & \multicolumn{1}{l|}{25}  & 0.63        & 0.64        & 0.64        & 0.64         \\
\multicolumn{1}{l|}{}                         & \multicolumn{1}{l|}{50}  & 0.65        & 0.66        & 0.66        & 0.66         \\
\multicolumn{1}{l|}{Records}                         & \multicolumn{1}{l|}{75}  & 0.69        & 0.70         & 0.70         & 0.70          \\
\multicolumn{1}{l|}{}                         & \multicolumn{1}{l|}{100} & 0.72        & 0.72        & 0.72        & 0.73        
\end{tabular}
\label{tab:table5}
\end{table}

\normalsize
It should be noted that the accuracy and F-measure of the algorithms do not significantly change when we down-sample the Gisette dataset into bytes.

\section{Conclusion and Future Works}
In this paper, an accurate and ultra-low power classifier was proposed. Its accuracy is comparable to state-of-the-art and traditional classifiers, and its time-complexity is linear. It was shown that it can run on ultra-low power devices, such as Arduino Uno, and training with 100 records and 100 features took 195 milliseconds. Applying the model on this microcontroller on a dataset with 100 records and 100 features took less than a millisecond. On a laptop, the runtime of SEFR was better than all of the other classifiers on binary datasets. As it is shown, the energy consumption of SEFR is such that it can be run on a device with very limited resources.

In future works, we aim to apply it in different domains and applications, such as image recognition, pose estimation, and other vision algorithms which utilize large amounts of battery from the device. Furthermore, we will develop a ultra-low power regression algorithm based on SEFR to enable microcontrollers run both train and test on-device.

\end{document}